\documentclass[sigconf,natbib=true,anonymous=false]{acmart}
\usepackage{balance}
\AtBeginDocument{%
  }

\copyrightyear{2026}
\acmYear{2026}
\setcopyright{cc}
\setcctype{by}
\acmConference[CIKM '26]{Proceedings of the 35th ACM International Conference on Information and Knowledge Management}{November 07--11, 2026}{Rome, Italy}
\acmBooktitle{Proceedings of the 35th ACM International Conference on Information and Knowledge Management (CIKM '26), November 07--11, 2026, Rome, Italy}
\acmDOI{10.1145/3799682.3840044}
\acmISBN{979-8-4007-2539-5/2026/11}

\begin{document}

\title{CG-Probes: Recovering Guardrail Directions from Patient Query Embeddings}

\author{Marko Řeháček}
\correspondingauthor
\orcid{0000-0003-2770-1299}
\email{rehacek@mail.muni.cz}
\affiliation{%
  \department{Faculty of Informatics}
  \institution{Masaryk University}
  \city{Brno}
  \country{Czech Republic}
}

\author{Vítězslav Dušek}
\authornote{Clinical oncologists who co-designed the axes and provided golden labels.}
\orcid{0000-0002-9875-5502}
\affiliation{
  \department{Dept.\ of Pediatric Oncology}
  \institution{University Hospital Brno}
  \city{Brno}
  \country{Czech Republic}
}
\additionalaffiliation{
  \department{Faculty of Medicine}
  \institution{Masaryk University}
  \city{Brno}
  \country{Czech Republic}
}
\additionalaffiliation{
  \institution{Masaryk Memorial Cancer Institute}
  \city{Brno}
  \country{Czech Republic}
}

\author{Martin Rusinko}
\orcid{0009-0001-0190-6344}
\affiliation{
  \institution{Masaryk Memorial Cancer Institute}
  \city{Brno}
  \country{Czech Republic}
}
\authornotemark[1]

\author{Vít Nováček}
\orcid{0000-0003-4687-6043}
\affiliation{
  \department{Faculty of Informatics}
  \institution{Masaryk University}
  \city{Brno}
  \country{Czech Republic}
}
\authornotemark[3]

\renewcommand{\shortauthors}{Marko Řeháček, Vítězslav Dušek, Martin Rusinko, and Vít Nováček}

\begin{abstract}
Patient-facing AI assistants promise valuable support to patients, but incoming queries can pose medical risks.
To create guardrails, we work with oncologists to define three ordinal risk axes: Medical Urgency, Psychological Urgency, and Topic Sensitivity. We propose Clinical Guardrail Probes (CG-Probes) to measure the risks from query embeddings.
We probe for each axis in the normalized embedding space of frozen embedders via the difference-in-means method, treating each axis as a potential linear direction.
To train the probes, we cluster 79,658 Czech oncology search queries with BERTopic and use these clusters to generate pairs of queries with contrastive risk levels via few-shot prompting.
We evaluate the approach on 200 queries (90 real, 110 synthetic), each graded by two oncologists, against two open-weight LLMs and a frontier LLM.
We find that urgency-based axes are recoverable as linear directions, and the probes are competitive with open-weight LLMs (no significant differences in quadratic-weighted $\kappa$) at a fraction of the latency.
Each axis yields a scalar score that clinicians can inspect and use to set escalation thresholds.
The pipeline requires only search logs, axis definitions, and black-box access to the embedding model, suggesting transferability across healthcare domains.
Robust validation on new queries and axes remains future work.
\end{abstract}



\maketitle

\section{Introduction}
Patient-facing AI assistants could improve access to health
information, but hospitals hesitate to deploy them because incoming queries range from
benign logistics (\emph{``where do I park?''}) through symptom
reports (\emph{``fever of 39\,°C after chemotherapy''}) to statements of
suicidal intent (\emph{``I no longer want to live''}).
This variation in risk makes clinician oversight and proper guardrails essential~\cite{wong_esmo_2025}.

Existing guardrail mechanisms offer three options. Relying on the
safety alignment of an LLM provides general harmlessness but cannot provide graded triage. Frontier LLMs undertriage specific emergency types and are inconsistent in distinguishing intermediate grades, agreeing with clinicians only at the extremes~\cite{mcbain_evaluation_2025,ramaswamy_chatgpt_2026}. Triage behavior varies across models and even across versions of the
same model~\cite{chen_how_2024, lee_performance_2025}. An LLM judge~\cite{inan_llama_2023, markov_holistic_2023} invoked before answering decouples safety from generation, but it adds latency and cost to every query, offers no inspectable decision boundary, and is unstable under model updates.
Rule-based filters~\cite{rebedea_nemo_2023} are both inspectable and auditable, but brittle.
No option simultaneously offers low latency, inspectability, auditability, and graded risk.

We therefore investigate a fourth option: defining medical risks as ordinal axes and testing whether each corresponds to a fixed direction in the L2-normalized space of frozen query embeddings. 
This setup is challenging because embedding models are usually optimized for retrieval, rewarding topic similarity~\cite{weller_nevir_2024}.
At the same time, decoder-only embedders initialized from autoregressive LLMs may retain clinical information from pretraining, and be prompted to recall it~\cite{singhal_large_2023, behnamghader_llm2vec_2024, peng_answer_2024,tao_llms_2026}.
A single dot product between a query embedding and a safety direction can then yield a scalar risk score. We ask: \emph{which clinician-defined risk constructs are linearly represented in frozen query embeddings of decoder-only embedders, and are the resulting scores accurate enough to serve as guardrails in a patient-facing setting?}
We investigate this in oncology, on three axes
(\S\ref{sec:axes}): Medical Urgency (MU), Psychological Urgency (PU), and Topic Sensitivity (TS).

\paragraph{\textbf{Our contributions are:}} (1)~an adaptation of difference-in-means probing to frozen, L2-normalized query embeddings, with synthesis of contrastive pairs from search logs to achieve within-pair topic cancellation; (2)~a clinician-labeled benchmark of 200 oncology queries, each graded by two oncologists, alongside a probe training pipeline; all released in an accompanying repository\footnote{Code repository containing the benchmark, probe training pipeline, and reports: \url{https://github.com/mrehacek/cg-probes}};
(3)~an empirical demonstration of which axes are linearly recoverable across four embedders, with the effects of instruction-conditioning analyzed;
and (4)~the feasibility of low-latency probe-based safety gating, with probes achieving competitive performance with an open-weight guardrail LLM (no significant QWK difference) at ${\sim}30\times$ higher sustained throughput and with no generated tokens.

\section{Related Work}
The idea that a concept can be represented as a direction in embedding space dates back to word embeddings: simple vector arithmetic was initially shown to solve word analogies~\cite{mikolov_linguistic_2013}, while difference-in-means (DiM) projections over contrastive sets were used to isolate bias directions~\cite{bolukbasi_man_2016}.
Recent interpretability work still applies these principles to autoregressive LLMs by \emph{probing} (i.e., reading from residual streams) and \emph{steering} (i.e., adding concept vectors during generation to shift model output) along linear directions~\cite{zou_representation_2025}.
The concept vectors can be computed as mean differences over contrastive pairs~\cite{rimsky_steering_2024}---requiring only a written description of the target concept~\cite{chen_persona_2025}---and DiM probes rival trained classifiers~\cite{marks_geometry_2024}. The probing requires white-box access to internal activations. Unlike these approaches, we work with the output vector of a frozen embedder, and target ordinal clinical constructs. Established probing methodologies and validation frameworks we use include diagnostic classifiers~\cite{alain_understanding_2016}, sentence-level evaluation tasks~\cite{conneau_what_2018}, control tasks~\cite{hewitt_designing_2019}, and documented pitfalls~\cite{belinkov_probing_2022}.

A prominent application of probing/steering is enforcing safe LLM behavior. High-level semantic information was found to be consistently residing in low-dimensional subspaces across open-weight models~\cite{saglam_large_2025}, including safety-related behaviors~\cite{wang_refusal_2026}, such as refusal, causally mediated by a single DiM direction~\cite{arditi_refusal_2024}.
More generally, safety constructs can be recovered as linear directions from a few hundred contrastive examples across multiple model families~\cite{arditi_refusal_2024, llorente-saguer_harmful_2026}. These findings suggest that clinical safety constructs might likewise be recoverable. The question is \emph{what} constructs.

Our construct design draws on two decades of work on health information needs. Early studies used search logs~\cite{spink_study_2004}, documented how medical
concerns escalate during search sessions~\cite{white_cyberchondria_2009}, and modeled intent in exploratory health search~\cite{cartright_intentions_2011}, a line continued by query-intent understanding~\cite{zhang_query_2020}. On the clinical side, patient questions have been coded into information-need taxonomies: triage models for portal messages that occasionally carry life-threatening symptoms~\cite{cronin_automated_2015},
coding schemes that combine query subject with question
type~\cite{boot_classifying_2010}, and resource-oriented taxonomies~\cite{roberts_resource_2016}.
These works classify query intent and risk descriptively with supervised classifiers, assigning each query a nominal category. Patient queries, however, routinely carry several kinds of risk at once, e.g. a report of a somatic symptom framed by distress, on a sensitive topic. We model risk as several ordinal axes rather than one category.

\section{Clinical Axes}
\label{sec:axes}
We define three ordinal axes (grades 0--2), co-designed with oncologists through an iterative codebook process. Each grade has an operational definition, anchor
phrases, and disqualifying markers. Table~\ref{tab:axes} summarizes the definitions, and the complete rubrics are released with the code repository.

\textbf{\emph{Medical Urgency (MU)}} grades reported somatic issues---from no action needed, through physician
contact, to activation of emergency services---inspired by CTCAE \cite{us_department_of_health_and_human_services_common_2025} and PRO-CTCAE~\cite{basch_development_2014} adverse-event grading systems and UKONS oncology triage ~\cite{uk_oncology_nursing_society_oncologyhaematology_2023}. Annotators grade only explicit text reports: a bare ``diarrhea'' is MU0 even when immunotherapy-induced
colitis is possible.

\textbf{\emph{Psychological Urgency (PU)}} grades a distress-to-suicidality
continuum~\cite{posner_columbiasuicide_2011} on the expressed thought, not
its phrasing: signs of distress raise the grade to PU1, and a calmly worded \emph{``I no longer want to live''} is PU2, while dramatic wording alone raises no grade.

\textbf{\emph{Topic Sensitivity (TS)}} grades the inherent sensitivity of the subject matter, following NCCN distress-management
guidelines~\cite{national_comprehensive_cancer_network_nccn_2024}, explicitly ignoring tone: a simple appointment request is TS0, and \emph{``will I lose leg''} is TS2, without verbalized distress. Unlike MU and PU, which grade what a query reports, TS grades the subject matter itself; it is therefore expected to correlate with topic by construction.

\begin{table}[t]
\caption{Clinical axes with shortened definitions and example queries. Complete rubrics are in the released code repository.}
\label{tab:axes}
\vspace{-7pt}
\footnotesize
\begin{tabular}{@{}cp{6.9cm}@{}}
\midrule
\multicolumn{2}{@{}l}{\textbf{Medical Urgency (MU)} -- time to clinical action} \\
0 & benign, no urgent somatic issue reported -- \emph{``I have a cough.''} \\
1 & physician contact required, emergency not excludable -- \emph{``Diarrhea six times a day, two days after chemotherapy.''} \\
2 & somatic emergency, activate EMS -- \emph{``My face is swollen can't breathe.''} \\
\midrule
\multicolumn{2}{@{}l}{\textbf{Psychological Urgency (PU)} -- expressed distress$\to$suicidality} \\
0 & no expressed distress -- \emph{``What are the side effects of radiotherapy?''} \\
1 & distress or existential doubt, no suicidal intent -- \emph{``I cry every day since the diagnosis and can't cope.''} \\
2 & suicidal intent, expert intervention required -- \emph{``I no longer want to live.''} \\
\midrule
\multicolumn{2}{@{}l}{\textbf{Topic Sensitivity (TS)} -- intrinsic topic load, tone ignored} \\
0 & administrative or logistical, regardless of referenced topic -- \emph{``Where do I go for screening?''} \\
1 & illness and treatment topics without lasting emotional or existential load -- \emph{``What does HER2-positive mean?''} \\
2 & high-load topics: prognosis, recurrence, end of life, body integrity, fertility, disclosure -- \emph{``How do I tell my children about the diagnosis?''} \\
\bottomrule
\end{tabular}
\end{table}

\section{Method}
\label{sec:method}

\subsection{\textbf{Embedding Models}}
We benchmark four embedders: Qwen3-Embedding-8B (4096d) \cite{qwen_team_qwenqwen3-embedding-8b_2026},
Harrier-oss-27b (5376d) \cite{microsoft_research_harrier-oss_2026}, 
text-embedding-3-large (3072d) \cite{openai_text_2024}, 
gemini-embedding-2 (3072d) \cite{google_ai_for_developers_gemini_2026}; all selected because they support Czech.
All vectors are L2-normalized; open-weight models are not fine-tuned and use last-token pooling. For instruction-following embedders we compare four conditioning modes, applied
identically to training pairs and evaluation queries: \emph{no instruction}, a single \emph{generic} instruction, a \emph{per-axis} instruction describing the axis, and a \emph{per-axis$^{+}$} variant omitting negative clauses.
One probe is trained per axis, per embedder, per instruction mode (Table \ref{tab:cells}).

\subsection{\textbf{Probe Design}}
We adapt difference-in-means probing, used to extract concept directions from word and sentence embeddings~\cite{bolukbasi_man_2016} and residual-stream activations~\cite{arditi_refusal_2024,chen_persona_2025}, to $L_2$-normalized contrastive embeddings. 
We estimate each probe from $N$ contrastive pairs $\mathcal{P}$ (\S\ref{sec:data}), pairing each baseline anchor $x_a$ with a higher-grade variant $x_v$ from the same semantic topic \eqref{eq:probe_direction}.
\begingroup
\setlength{\abovedisplayskip}{4pt}
\setlength{\belowdisplayskip}{4pt}
\begin{equation}\label{eq:probe_direction}
w = \frac{1}{N} \sum_{(x_v, x_a) \in \mathcal{P}} \bigl( \mathrm{emb}(x_v) - \mathrm{emb}(x_a) \bigr), \qquad \hat{w} = w / \lVert w \rVert_2 .
\end{equation}
\endgroup
Each pair shares the same topic, so their embedding difference suppresses topic-related differences. Averaging the differences across pairs then reinforces the direction associated with higher grades. Systematic differences between anchors and their variants, such as length, register, or generation style, could also accumulate in $w$. We mitigate their effects by generating variants from exemplars in the anchor's topic cluster and prompting them to preserve the anchor's length and register (\S\ref{sec:data}).

\subsection{\textbf{Controlling Topic Confounds}}
Training a probe directly on ${\{query, grade\}}$ labels can leak topic identity: a cluster
about liver metastases separates trivially from one about parking on every
axis, allowing a probe to predict grade from the topic
rather than the target
construct~\cite{hewitt_designing_2019,belinkov_probing_2022}.
We address this in two ways. First, each anchor is paired with a higher-grade variant from the same topic, so topic identity alone cannot determine the grade. Second, we split the training, validation, and evaluation sets by topic cluster---each cluster is assigned to exactly one split. Pairing queries within the same topic also avoids confounding grade with topic when estimating the direction: independently sampled grade classes can have different topic distributions, and their mean difference may therefore reflect topic variation. This is particularly problematic in anisotropic embedding spaces, where variance can concentrate in a few directions~\cite{ethayarajh_how_2019}.

\subsection{\textbf{Baselines and Metrics}}
We compare the probes with zero-shot classification by two open-weight
LLMs realistically deployable in hospital settings, 
gpt-oss-safeguard-20b \cite{openai_gpt-oss-safeguard_2025} and gpt-oss-120b \cite{openai_gpt-oss-120b_2025}. Both are prompted
with the grade definitions and twelve boundary examples, and a single inference pass grades all three axes.
The safeguard model's reasoning effort is set to low, based on a sweep showing no improvement at higher effort while increasing latency (low effort takes ${\sim}3$ seconds per query; see \S\ref{sec:results}).
We also report a frontier model as an upper reference, gpt-5.4~\cite{openai_openai_2026} with medium reasoning. Our primary metric is quadratic-weighted
$\kappa$~\cite{cohen_weighted_1968} (QWK) with 95\% bootstrap CIs (1,000 resamples), reported against the inter-oncologist
ceiling (\S\ref{sec:data}). We additionally report macro-F1 and, for the probes, AUROC for detecting grade-2 queries; the LLM baselines produce discrete grades and therefore do not provide a ranking score.

\section{Data, Pipeline, and Benchmark}
\label{sec:data}

\subsection{Data and Clustering}
We collected 79{,}658 de-duplicated Czech patient search queries from a comprehensive cancer center website. BERTopic~\cite{grootendorst_bertopic_2022} (over
Qwen3-Embedding-4B embeddings~\cite{qwen_team_qwenqwen3-embedding-8b_2026}, with
UMAP~\cite{mcinnes_umap_2020} and HDBSCAN~\cite{mcinnes_hdbscan_2017})
yields $1{,}889$ superclusters. We then use gpt-5.4-mini \cite{openai_openai_2026} to remove irrelevant clusters, selecting 861 clusters for training and 134 for benchmark.

\subsection{Evaluation Pool}
The pool is built in four steps. Model families and configurations were chosen for Czech support and after manual comparison of generation quality.

\textbf{\emph{(1) Pick}}: gpt-5.4 reads each
cluster's queries and selects queries likely to have higher grades on any axis,
assigning grades on all three axes (10{,}128 picks); non-clinical clusters are skipped and
navigational ones are subsampled.

\textbf{\emph{(2) Sample}}: a greedy sampler selects 400 queries to match target grade marginals, with a limit per cluster, preferring oncology-related queries and
high-confidence picks; the 290 real queries come from 134 clusters.

\textbf{\emph{(3) Verify}}: gpt-5.4-mini re-grades all queries seeing only
the query and the rubrics, and queries with labels disagreeing with the initial label from gpt-5.4 are labelled as ambiguous.

\textbf{\emph{(4) Backfill}}: MU2 and PU2 cells are filled with synthetic queries
generated from clusters containing MU1/PU1, confirmed at the target grade by an independent
verifier, and swapped in for the ambiguous MU0/PU0 picks.

The resulting pool contains 400 queries: 290 real and
110 synthetic (28\%).
Synthetic queries were reviewed by an oncologist for clinical plausibility and accepted with few exceptions. Query generation is grounded in topic clusters; approximately 65\% of MU2 and 97\% of PU2 queries are synthetic.
Per-axis grade marginals (0/1/2) are MU 191/129/80, PU 248/92/60, TS 42/277/81.

\subsection{Clinician Gold}
A subset of 200 queries (90 real, 110 synthetic) was graded
independently on all three axes by two oncologists.
We then jointly reviewed every disagreement to identify annotation errors or ambiguities in the codebook. MU and PU disagreements were resolved easily. For TS, 100 of 101 disagreements followed the same pattern regarding a specific boundary: whether acute somatic urgency implies a high-sensitivity topic. We resolved this by a systematic rule. TS2 denotes lasting emotional or existential load, while somatic urgency alone remains TS1. One unresolved TS item was dropped (TS $n{=}199$).
Inter-oncologist agreement~\cite{artstein_inter-coder_2008} sets the empirical ceiling for MU and PU at QWK 0.78 and 0.96, respectively.
The TS ceiling is 0.82 after adjudication, but we report TS descriptively and do not rank systems on this axis (\S\ref{sec:results}).

\subsection{Evaluation Protocol}
Primary evaluation uses $n=90$ real queries ($n=89$ for TS after dropping one unresolved item). The results of the pooled set are reported alongside. Synthetic queries are substantially easier for the LLM baselines (0.94--0.99 QWK on synthetic queries versus 0.64--0.89 on real queries). Probe performance is similar on synthetic and real queries (e.g., MU: 0.63 vs.\ 0.65). Probe configurations (embedder $\times$ instruction mode) are selected by QWK on the real set, introducing an optimistic selection bias; the LLM baselines involve no analogous selection. All 400 queries carry silver labels from gpt-5.4, which we use only for a larger-$N$ robustness check and emergency-recall analysis (\S\ref{sec:results}).

\subsection{Contrastive Pairs Generation}
\label{sec:contrastive_pairs_generation}
From the previously LLM-labeled $10{,}128$ queries we build an anchor
pool of $8{,}825$ queries, excluded from the evaluation set by query and by
cluster. Real logs contain almost no grade-2 safety queries (28 MU2 and
2 PU2, all reserved for evaluation), so grade-2 and PU1
training positives are synthesized as rewrites of real
anchors via few-shot prompting~\cite{brown_language_2020,wang_self-instruct_2023}. Two safeguards
protect against synthetic artifacts. First, the training generator
(gemini-3.5-flash~\cite{google_deepmind_gemini_2026}) is a different model family from the evaluation set's
selector (gpt-5.4~\cite{openai_openai_2026}), so separability that survives the mismatch is unlikely to reflect a generation style specific to either model. Second, independent verifier (gpt-5.4-mini~\cite{openai_openai_2026}), blind to the target grade, re-grades every variant, keeping only confirmed items
(94\% MU, 90\% PU). TS pairs are built entirely from real queries.

The resulting training sets contain 3{,}114 (MU), 3{,}158 (PU) and 2{,}116 (TS) graded items drawn from 460/478/733 clusters, split cluster-disjointly into
train/dev/test (1{,}730/619/765, 1{,}825/631/702, 1{,}319/384/413). The direction is estimated from the grade-2 items of the train split alone, so $N=427$ (MU), $395$ (PU) and $553$ (TS) in Eq.~\ref{eq:probe_direction}. Each grade-2 query is differenced against its paired grade-0 anchor where one exists, else against the grade-0 mean of its own cluster, else against the global grade-0 mean. A paired anchor is available for 79\% of MU and 54\% of PU grade-2 items, but for 0\% of TS; 82\% of TS items fall back to the global grade-0 mean (MU 0\%, PU 5\%). Topic cancellation is therefore not applied on TS.
Grade-1 queries do not enter $w$ and are used only to
tune the thresholds for emergency recall test (\S\ref{sec:emergency_recall}). 

\section{Results}
\label{sec:results}

\begin{figure}[t]
  \centering
  \captionsetup{skip=6pt}
  \includegraphics[width=\columnwidth]{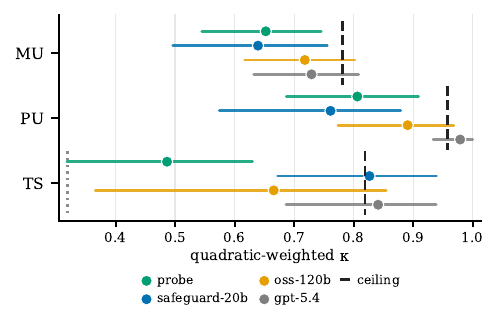}

  \vspace{2pt}
  {\footnotesize\setlength{\tabcolsep}{5pt}
  \begin{tabular}{lccccc}
    \toprule
    Axis & Ceiling & Probe & safeguard-20b & oss-120b & gpt-5.4\\
    \midrule
    MU & 0.78 & 0.65 & 0.64 & 0.72 & 0.73\\
    PU & 0.96 & 0.81 & 0.76 & 0.89 & 0.98\\
    TS & 0.82 & 0.58 & 0.83 & 0.67 & 0.84\\
    \bottomrule
  \end{tabular}}

  \caption{\textbf{Probe versus LLM guardrails, against the clinician gold and the
human ceiling.}
Quadratic-weighted $\kappa$ on real queries; bars are 95\% bootstrap CIs, the ceiling is inter-oncologist agreement on the same items. Probe = the 1-D difference-in-means direction at each
axis's best embedder $\times$ instruction cell (Table~\ref{tab:cells}) for MU and PU; for TS, the class-balanced linear head (0.58), as a single direction underfits TS (0.49, Table~\ref{tab:cells}; §6). The probe is statistically indistinguishable from both open-weight LLMs on MU and PU, and is
the weakest on TS, where we make no ranking (\S\ref{sec:results}).
}
\Description{Grouped bar chart comparing quadratic-weighted kappa
of the CG-Probe against three LLM baselines and the inter-oncologist
ceiling, for the three risk axes.}
  \label{fig:main}
\end{figure}

\begin{table}[t]
  \centering
  \captionsetup{position=bottom, skip=8pt}
  {\footnotesize\setlength{\tabcolsep}{5pt}
  \begin{tabular}{llccc}
    \toprule
    embedder & instruction & MU & PU & TS \\
    \midrule
    Qwen3-8B (4096d) & none & 0.34 & 0.67 & 0.38 \\
     & generic & 0.34 & \textbf{0.81} & 0.30 \\
     & per-axis & 0.61 & 0.67 & 0.37 \\
     & per-axis$^{+}$ & 0.60 & 0.58 & 0.35 \\
    \addlinespace
    Harrier-27B (5376d) & none & 0.30 & 0.62 & 0.36 \\
     & generic & 0.51 & 0.09 & 0.27 \\
     & per-axis & 0.62 & 0.45 & 0.29 \\
     & per-axis$^{+}$ & 0.62 & 0.47 & 0.29 \\
    \addlinespace
    text-emb-3-large (3072d) & --- & 0.25 & 0.51 & 0.42 \\
    \addlinespace
    gemini-emb-2 (3072d) & none & 0.21 & 0.61 & 0.45 \\
     & generic & 0.47 & 0.59 & \textbf{0.49} \\
     & per-axis & \textbf{0.65} & 0.64 & 0.26 \\
     & per-axis$^{+}$ & 0.63 & 0.49 & 0.24 \\
    \bottomrule
  \end{tabular}}
  \caption{Quadratic-weighted $\kappa$ of the
1-D DiM probe against the clinician gold, on the same real queries as
Fig.~\ref{fig:main}, for embedders $\times$ instruction modes.
  Instruction conditioning helps MU and is mixed for PU. No embedder or instruction recovers TS.
Per-axis$^{+}$ is the negation-free instruction variant;
text-embedding-3-large accepts no instruction.
Best cells reported in Fig.~\ref{fig:main} are in bold, the selection bias is noted in \S\ref{sec:data}.}
  \label{tab:cells}
\end{table}

\paragraph{\textbf{Urgency axes are competitive with LLMs}}
Figure~\ref{fig:main} summarizes the main results; Table~\ref{tab:cells} reports the tuning.

On MU and PU, the probe shows no significant QWK difference from the open-weight
LLMs: paired on the same 89 items, $\Delta$QWK is $+0.02$/$+0.05$ (MU/PU)
against safeguard-20b and $-0.06$/$-0.08$ against gpt-oss-120b, with every
95\% bootstrap CI including zero. We claim competitiveness, not
equivalence: a TOST analysis \cite{schuirmann_comparison_1987} certifies equivalence only within
${\pm}0.18$ at $n{=}89$, so tighter margins require a larger evaluation.
Only the frontier model separates clearly, and only on PU (0.98, at the human
ceiling). Conclusions are stable across label sets: moving from silver to
gold shifts MU/PU by ${\leq}0.04$ QWK, while silver over-credited the 1-D
TS probe by $0.15$.

Instruction-conditioning behaves differently per axis (Table~\ref{tab:cells}). A per-axis instruction lifts MU (0.21--0.51 raw or generic, 0.60--0.65 per-axis), PU is mixed and best under a single generic instruction (0.81), and TS is degraded by conditioning (gemini: 0.49 generic vs. 0.26 per-axis).

\paragraph{\textbf{ Topic Sensitivity recovers a topic direction, not a sensitivity direction}}
The 1-D TS probe is the weakest (0.49 QWK). A class-balanced linear
head reaches 0.58 and MLP provides no further gain, suggesting that TS does not have a single direction. Topic membership explains $\eta^2{=}0.81$ of the TS projection variance, compared with 0.12/0.22 for MU/PU. 
By construction, TS lacks topic-matched pairs; 82\% of positives are differenced against a global grade-0 mean (\S\ref{sec:contrastive_pairs_generation}) representing administrative text (TS0). The probe thus learns a generic clinical vs. administrative axis (the easy TS0/TS1 boundary) rather than the subtle emotional-load distinction between TS1 and TS2.
In counterfactual pairs, changing the grade moves the TS projection only 1.9$\times$ as much as a topic-preserving paraphrase, with the movement in the expected direction in 78\% of cases. We therefore do not rank TS, and read this as a limitation of the pair construction rather than evidence no TS direction exists.

\paragraph{\textbf{The MU probe misses no emergencies, at 19.5\% escalation of benign queries}}
\label{sec:emergency_recall}
Deployed at the thresholds $(t_1,t_2)=(-0.043,\,0.082)$ on the unit-normalized projection, tuned once to maximize macro-F1. Scores below $t_1$ map to MU0, above $t_2$ to MU2, and intermediate to MU1. An emergency is missed only when a system grades it MU0, as higher grades should trigger clinical action.

On 28 real, silver-labeled clear somatic emergencies, the probe missed none, whereas safeguard-20b missed 8. The four cases that the probe downgraded were still graded MU1. The clinician-labeled subset shows the same pattern: the probe missed 0 of 9 gold MU2 queries, compared with 1 of 9 for safeguard-20b. Increasing the safeguard model’s reasoning effort made it both slower and less safe: missed emergencies increased from 8 at low effort (2.9\,s) to 12 at medium effort (7.0\,s). This recall is bought with 19.5\% of the 133 benign real queries being escalated to at least MU1, where safeguard-20b
escalates none. The probe emits a continuous score (AUROC 0.89 MU, 0.93 PU), so $t_1$ and $t_2$ are free parameters a hospital can move along that trade-off. The LLM baselines have no such control.

\paragraph{\textbf{Avoiding LLM generation buys a speedup}}
Probes require only dot products after a single batched encoder pass.
Although our embedders are themselves large models, replacing
safeguard-20b's $\sim360$ decode steps (2.9\,s on an NVIDIA L40S) with
one encoder pass is $\sim10^{2}$--$10^{3}\times$ cheaper with $\sim30\times$ sustained throughput (measured end-to-end).
This advantage comes from avoiding autoregressive generation.
Probes can therefore serve as a cheap first layer for safety screening,
and can reuse an embedding model already deployed for retrieval in a RAG-based system ~\cite{lewis_retrieval-augmented_2020}.

\paragraph{\textbf{Controls argue against generator-specific artifacts}}
We test the main design choices and whether the learned directions reflect the target grades rather than topic or generation artifacts.

\emph{(1) Probe design.} A class-balanced linear head and an MLP do not improve TS beyond the 1-D probe. Instruction-conditioning helps MU/PU but degrades TS (Table \ref{tab:cells}).
Removing negative clauses (disqualifying rules) from TS instruction does not help, which suggests conditioning suppresses topic signal that TS depends on.

\emph{(2) Generalization.} The probes generalize to real golden-labeled queries, despite being trained on contrastive pairs whose positive examples are partly synthetic.

\emph{(3) Control task.} Following~\cite{hewitt_designing_2019}, we refit each probe with grade labels randomly shuffled across queries, so that the labels contain no information about the target construct; macro-F1 drops by 0.32/0.49/0.37 (MU/PU/TS) to chance.

\emph{(4) Counterfactual test.} In topic-matched pairs, changing only the MU/PU grade moves the probe projection 3.7$\times$/6.2$\times$ more than paraphrasing the query without changing its grade. The projection moves in the expected direction in 94--99\% of cases.

\emph{(5) Direction separation.} Probes are moderately correlated, as expected: MU$\leftrightarrow_{\mathrm{cos}}$PU (0.40--0.60), TS$\leftrightarrow_{\mathrm{cos}}$others (0.13--0.50).

\emph{(6) Synthetic artifacts.} Using different model families for training generation and evaluation selection, and independently verifying generated variants reduces the risk (\S\ref{sec:data}).

\section{Limitations}
\label{sec:limitations}

\textbf{Statistical power.} The gold contains 90 real queries. This supports our claim of competitiveness on the urgency axes, but not equivalence: at this $n$, TOST certifies equivalence within $\pm0.18$ QWK.

\textbf{Synthetic safety cells.} Because high-grade queries are rare in search logs, approximately 65\% of MU2 and 97\% of PU2 queries in the evaluation pool are synthetic. Generalization to real high-risk queries, particularly for PU2, is therefore least certain.
At the same time, the synthetic generation gives us a practical way to train the guardrails; and deploying sufficiently reliable guardrails could in turn enable safer collection of additional real high-risk queries.
The emergency-recall analysis nevertheless uses 28 real, silver-labeled MU2 queries, with a clinician-labeled cross-check of $n=9$. Synthetic artifacts cannot be ruled out, but we use independent verification and validity checks to reduce this risk (\S\ref{sec:results}).

\textbf{TS convention-dependence.} The reconciled TS gold encodes one
adjudicated coding convention (\S\ref{sec:data}), so we report TS
descriptively and make no TS system ranking.

\textbf{Single language, site, and register.} We use Czech single-turn search queries from one oncology center. While the pipeline does not use language- or site-specific components, cross-language and cross-site
generalization remains untested, and conversational input (longer, multi-turn,
shifted register) requires revalidation (\S\ref{sec:method}).

\section{Conclusion}
\label{sec:conclusion}
We show that certain clinical risk axes are linearly represented in frozen query embeddings, while others are not.
Medical Urgency (MU) and Psychological Urgency (PU) can be recovered as single linear directions, producing probes that perform competitively with open-weight LLMs. Avoiding autoregressive generation results in ${\sim}30\times$ higher throughput with zero decoding latency. Furthermore, the MU probe misses fewer emergencies than the guardrail LLM when used as a triage mechanism. In contrast, Topic Sensitivity is not recovered by a single direction: because TS grades subject matter, topic-matched pairs cannot be built at our cluster granularity.
Future work must evaluate real-world reliability and test additional risk axes.
More broadly, clinical guardrail probes enable safer collection of rare high-risk queries needed to advance safety of the assistants.
Finally, because a probing layer can be frozen and kept separate from the response generator, it can be audited independently as the underlying model evolves, making it a practical safety component for patient-facing AI systems~\cite{us_food_and_drug_administration_marketing_2025}.

\begin{acks}
This work has been funded by the Ministry of Health of the CR under the EMPOWER project (grant no. NW25-09-00465) and Technology Agency of the Czech Republic under the DecideHealth project (grant no. TQ12000018). Computational resources were provided by the e-INFRA CZ project (ID:90254), supported by the Ministry of Education, Youth and Sports of the Czech Republic.
\end{acks}

\section*{Ethics and Privacy Statement}
Search logs were used under an agreement with Masaryk Memorial Cancer Institute as part of the EMPOWER project, approved by the hospital's IRB. The agreement permits publication of de-identified patient data. Search log entries were de-identified using a self-hosted LLM before processing by any third-party API.
The main identified risk is that a probe could miss medically urgent queries. Beyond missed urgent queries, false positives can impose load on clinicians if they must review them in a live system. The probes are not currently deployed to patients, and are intended to supplement, not replace, established clinical safety pathways; they are not presented as autonomous clinical decision-makers.

\section*{GenAI Usage Disclosure}
LLMs are a studied component of this work, not only a
writing aid. \textbf{In the pipeline}, LLMs were used to filter
irrelevant topic clusters and to verify synthetic items (gpt-5.4-mini),
to pick and label candidate queries and generate the
evaluation-set synthetic items (gpt-5.4), and to generate the training-set
synthetic pairs (gemini-3.5-flash); gpt-oss-120b and gpt-oss-safeguard-20b
are evaluated baselines and gpt-5.4 additionally serves as a frontier
reference. All such uses are described in
\S\ref{sec:method}--\ref{sec:data}, and the pipelines, prompts, and cached embeddings
are released with the benchmark. \textbf{In implementation}, Claude
(Anthropic) was used as a coding assistant to help implement the
data processing, training, and evaluation pipelines and to anonymize the
released repository; all generated code was reviewed and tested by the
authors. \textbf{In preparing the manuscript}, generative AI assistants
were used for copy-editing, LaTeX formatting, and structural suggestions.
All research design, analysis, results, and claims are the authors' own,
and the authors take full responsibility for the entire content.

\bibliographystyle{ACM-Reference-Format}
\balance
\bibliography{references}

\end{document}